\documentclass[10pt,twocolumn,letterpaper]{article}

\usepackage[pagenumbers]{cvpr} 

\usepackage{graphicx}
\usepackage{amsmath}
\usepackage{amssymb}
\usepackage{booktabs}
\usepackage{caption}
\usepackage{subcaption}

\usepackage[pagebackref,breaklinks,colorlinks]{hyperref}

\usepackage[capitalize]{cleveref}
\crefname{section}{Sec.}{Secs.}
\Crefname{section}{Section}{Sections}
\Crefname{table}{Table}{Tables}
\crefname{table}{Tab.}{Tabs.}

\begin{document}

\title{Auto-Platoon : Freight by example}

\author{Abhijay Singh\\
Maryland Applied Graduate Engineering\\
University of Marlyand\\
{\tt\small abhijay@umd.edu}
\and
Tharun Puthanveettil\\
Maryland Applied Graduate Engineering\\
University of Marlyand\\
{\tt\small tvpian@umd.edu}
\and
Vinay Bukka\\
Maryland Applied Graduate Engineering\\
University of Marlyand\\
{\tt\small vinay06@umd.edu}
\and
Sameer Arjun S\\
Maryland Applied Graduate Engineering\\
University of Marlyand\\
{\tt\small ssarjun@umd.edu}
\and
Yashveer Jain\\
Maryland Applied Graduate Engineering\\
University of Marlyand\\
{\tt\small yashveer@umd.edu}
\thanks{\href{https://github.com/yashveerjain/Leader-Follower}{https://github.com/yashveerjain/Leader-Follower}}
}
\maketitle

\begin{abstract}
The work introduces a bio-inspired leader-follower system based on an innovative mechanism proposed as software latching that aims to improve collaboration and coordination between a leader agent and the associated autonomous followers. The system utilizes software latching to establish real-time communication and synchronization between the leader and followers. A layered architecture is proposed, encompassing perception, decision-making, and control modules. Challenges such as uncertainty, dynamic environments, and communication latency are addressed using Deep learning and real-time data processing pipelines. The follower robot is equipped with sensors and communication modules that enable it to track and trace the agent of interest or avoid obstacles. The followers track the leader and dynamically avoid obstacles while maintaining a safe distance from it. The experimental results demonstrate the proposed system's effectiveness, making it a promising solution for achieving success in tasks that demand multi-robot systems capable of navigating complex dynamic environments.
\end{abstract}

\section{Introduction}
\label{sec:intro}
This project addresses the problem of object detection, state estimation, and depth estimation using a custom-built algorithm to get the POSE values of the leader vehicle. Based on this data the path of the follower(s) is generated. Additionally, the followers are also added with dynamic obstacle avoidance capabilities to navigate their path towards the leader while maintaining a fixed following distance.

\subsection{Contribution - Software Latching}
In this study, we propose a software latching mechanism that ensures the safe and reliable engagement or disengagement of the autonomous mode of operation for the follower robot. It is crucial that components of autonomous systems maintain appropriate mode transitions to guarantee the system's operational safety. Software latching ensures that the activation and deactivation of autonomous  behavior in the follower robot are intentional and safe.

When a software latch is engaged, it means that the follower robot is operating autonomously and relying on its sensors and algorithms to track and follow the leader. This typically occurs when certain conditions are met, such as the leader robot being detected and recognized by the follower robot's perception system. The software latch helps prevent accidental or erroneous activation or deactivation of autonomous driving functions. It requires intentional action or specific triggering conditions to enable or disable the autonomous mode. This intentional action can be initiated by the user or by a higher-level control system that supervises the leader-follower system. 

In an autonomous leader-follower system, software latching also includes communication mechanisms between the leader and follower robots. The leader robot transmits signals or commands to the follower robot, indicating that it should enter or exit the autonomous mode. These signals can act as triggers for the follower robot's software latch, enabling or disabling the autonomous behavior based on the leader's instructions.

Furthermore, software latching often includes fail-safe mechanisms to handle situations where the system encounters errors or is unable to maintain safe operation. For instance, if the autonomous leader-follower system detects a critical issue or if the environmental conditions deteriorate beyond the system's capabilities, it may trigger an automatic disengagement of the autonomous mode and hand over control back to the operator.

\section{Background}
\subsection{Bioinspired Duck Walking Behaviour}
\begin{figure}[h!]
    \centering
    \includegraphics[width=\linewidth]{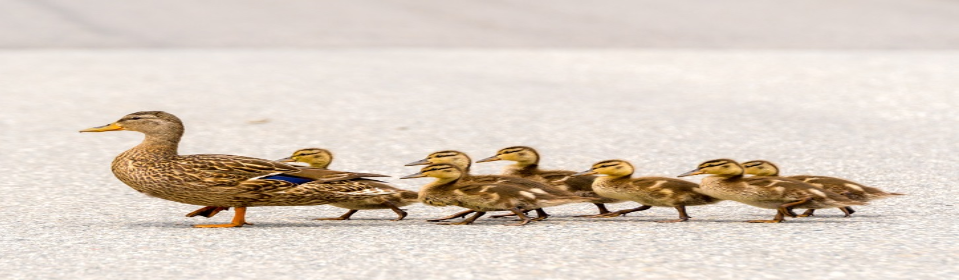}
    \caption{Bio-inspired duck behavior}
    \label{fig:bio-inspired-duck}
\end{figure}

The aforementioned technique of software latching, which is the foundation of the implementation in this study, is a bio-inspired mechanism from ducks. Ducks often exhibit a natural leader-follower behavior when walking in a group as shown in Fig.\ref{fig:bio-inspired-duck}. This algorithm can be applied to various areas, including swarm robotics, traffic control, and distributed systems.

The leader-follower algorithm is based on the idea of having a designated leader that guides the rest of the group, while the followers mimic the leader's actions and movements. In the case of ducks walking, the leader is typically the dominant or most experienced duck and the followers maintain a certain distance and direction relative to the leader.

When applied to swarm robotics, the leader-follower algorithm can be used to coordinate a group of robots. One robot is designated as the leader, and the remaining robots follow its movements. The leader may have access to more information, such as a global objective or environmental cues, which it communicates to the followers. The followers then adjust their behavior accordingly to maintain formation and achieve the collective goal.

\subsection{YOLOv7}

Object detection is a critical task in autonomous driving, as it enables vehicles to detect and track other vehicles, pedestrians, and obstacles in real-time to make informed decisions and avoid collisions. YOLO\cite{wang2022yolov7} is a popular deep learning-based approach for object detection that has demonstrated state-of-the-art performance in several benchmarks. The YOLO algorithm works by dividing an input image into a grid of cells and predicting object bounding boxes and their associated class probabilities for each cell. YOLO uses a single convolutional neural network (CNN) to simultaneously predict the class probabilities and bounding boxes of multiple objects in the image. The YOLO network architecture consists of 24 convolutional layers followed by 2 fully connected layers. The network takes an input image and outputs a grid of cells, each containing information about the predicted bounding boxes and their associated class probabilities. Each cell predicts a fixed number of bounding boxes, each of which has 5 values: the (x, y) coordinates of the box center, the width and height of the box, and the objectness score, which indicates the likelihood that the box contains an object. During training, YOLO minimizes a loss function that combines localization loss and classification loss. The localization loss measures the error between the predicted bounding boxes and the ground truth boxes, while the classification loss measures the error in predicting the class probabilities. YOLO also uses a technique called anchor boxes to improve the accuracy of bounding box predictions. Anchor boxes are pre-defined shapes of different sizes and aspect ratios that are used to predict the bounding boxes instead of directly predicting the width and height.
\begin{figure*}[h!]
    \centering
    \includegraphics[width=1\textwidth]{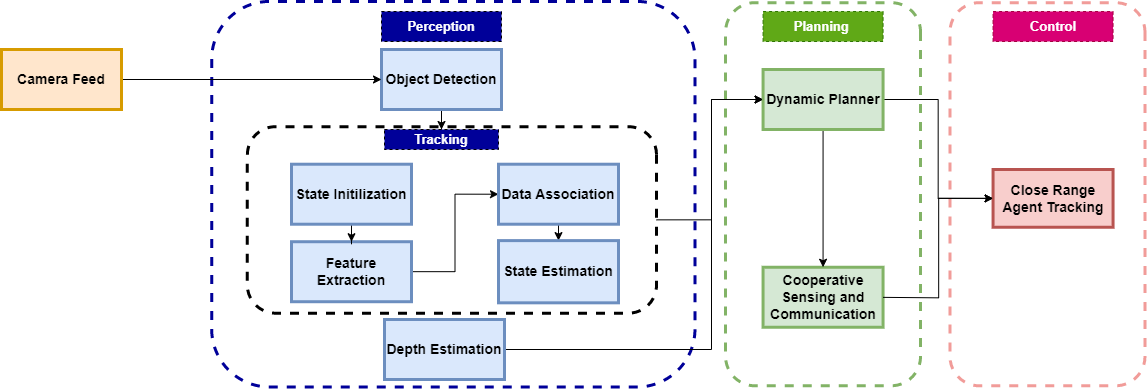}
    \caption{Process Flow of the Auto Platoon system}
    \label{fig:my_label}
\end{figure*}

\subsection{MiDaS}
MiDaS\cite{Ranftl2022} is a deep learning-based approach for depth estimation from a single monocular image. The MiDaS architecture consists of a multi-scale encoder-decoder network with skip connections. The encoder network consists of multiple convolutional layers that downsample the input image to extract features at multiple scales. The decoder network consists of several upsampling layers that reconstruct the output depth map from the encoded features. Skip connections are used to connect the encoder and decoder networks at multiple scales, allowing the decoder to access features from different levels of the encoder. To train the MiDaS network, a novel training loss called MiDaS loss is used. The MiDaS loss is a combination of a depth consistency loss and a gradient similarity loss. The depth consistency loss measures the difference between the predicted depth map and the ground truth depth map, while the gradient similarity loss encourages the predicted depth map to have similar gradients to the input image. By using the MiDaS loss, the MiDaS network can learn to estimate depth maps that are not only accurate but also visually plausible. One of the key strengths of MiDaS is its robust performance in challenging scenarios such as low-light conditions, occlusions, and reflections. This is achieved by incorporating a confidence map into the depth estimation process, which allows MiDaS to assign lower confidence to regions with low texture or high uncertainty. The confidence map is computed based on the depth consistency between neighbouring pixels, which helps MiDaS to handle occlusions and reflections more effectively. MiDaS has been extensively evaluated on several benchmark datasets, including the KITTI and Make3D datasets. It has consistently achieved state-of-the-art performance in terms of accuracy and robustness.

\section{Methodology}
The methodology for implementing the multi-agent leader-follower system involves the following steps:

\subsection{Detection}
The first step is to detect the object of interest in the input data, from the video feed.  Object detection algorithms are used to identify and localize the object within the given scene. This component helps initialize the tracking process. 
\subsubsection{Model}
A custom-trained Yolov7 model from the Yolo family is used to detect the target object of interest(robot) by the follower(agent/robot).
\subsubsection{Dataset}
\begin{figure}[h!]
    \centering \includegraphics[width=0.65\columnwidth]{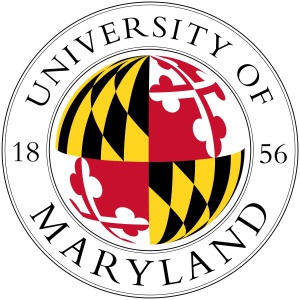}
    \caption{Target object used for training the Yolo-V7 model.}
    \label{fig:dataset}
\end{figure}

\begin{figure}[h!]
    \centering \includegraphics[width=0.65\columnwidth]{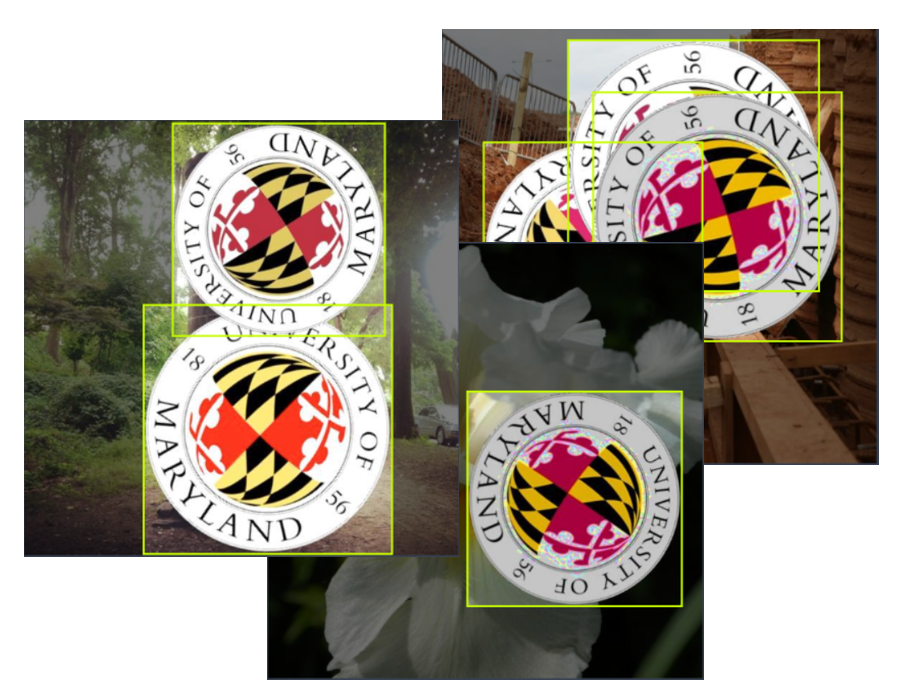}
    \caption{Annotated sample from the dataset}
    \label{fig:annotated_synthetic_data_sample}
\end{figure}

The dataset used for the custom model consisted of the University of Maryland logo (Fig.\ref{fig:dataset}). Synthetic data generation was used to create the annotated dataset required for training the object detection model (Fig.\ref{fig:annotated_synthetic_data_sample}). The number of training, validation, and testing samples were 4200, 400, and 200 respectively.

\subsection{Tracking}
\begin{itemize}
    \item State Initialization: Once the object is detected, an initialization step is performed to create a tracking representation or model of the object \cite{8909903}. This could involve extracting relevant features or creating a bounding box around the object to establish its initial position. We initialize the tracking id to the new object detected and assign a feature and bounding box, which then use been down the line for tracking. 
    
    \item Feature Extraction: 
    Feature extraction is a technique employed to capture unique and pertinent attributes of an object, which can then be utilized for tracking purposes. These features encompass various aspects such as color, texture, shape, or motion information. The extraction process aims to capture distinctive characteristics that allow for the effective representation of the object, enabling differentiation from the background or other objects.
    
    In our specific case, we utilized a convolutional neural network (CNN) model called MobileNet V2 as the feature extractor \cite{2019arXiv190608172L}. This CNN model was employed to extract feature embeddings from the detected object. These feature embeddings serve as compact representations that encode the essential information of the object, facilitating subsequent tracking and analysis tasks.
    
    \item Data Association: 
    Data association involves establishing a connection between an initially represented object and the corresponding object identified in subsequent frames. This process helps maintain the object's identity during tracking, even in the presence of occlusions or brief disappearances. To compare the features of the detected object across frames, cosine similarity is utilized as a measurement.
    
    In our scenario, we have set the matching threshold to 0.65. When we obtain the best matching feature, we average it with the previous feature. By shifting the centroid of the object's feature vector space to the optimal position, we can accommodate more variations of the same object, thereby enhancing the accuracy of the matching process. If an object is not detected for a certain period, despite having an assigned tracking ID, the object ID is eventually removed from the memory.

    \begin{figure}
        \centering
        \begin{subfigure}{0.65\textwidth}
             \includegraphics[width=0.6\textwidth]{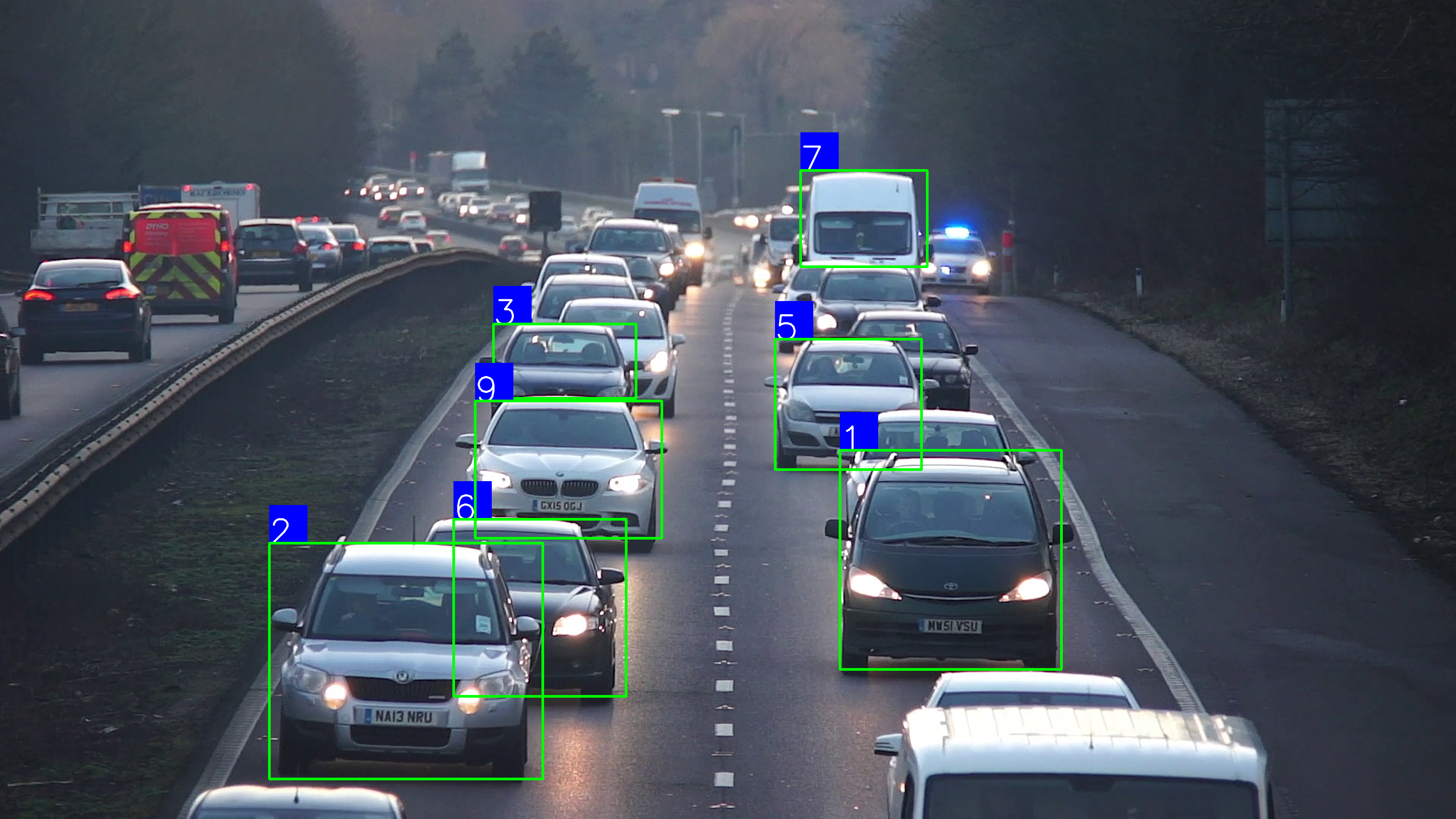}
        \end{subfigure}
        \begin{subfigure}{0.65\textwidth}
             \includegraphics[width=0.6\textwidth]{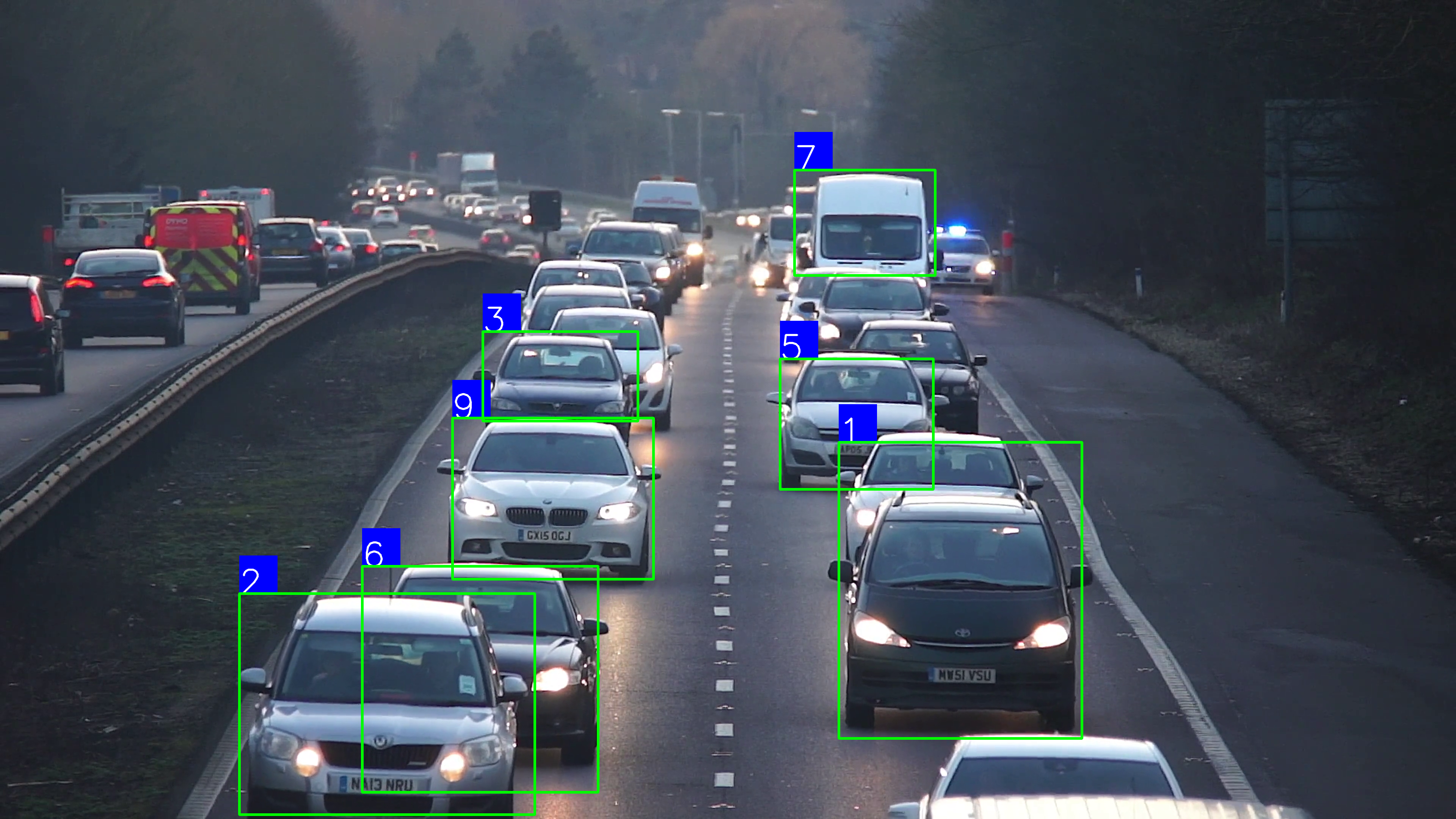}
        \end{subfigure}
        \begin{subfigure}{0.65\textwidth}
             \includegraphics[width=0.6\textwidth]{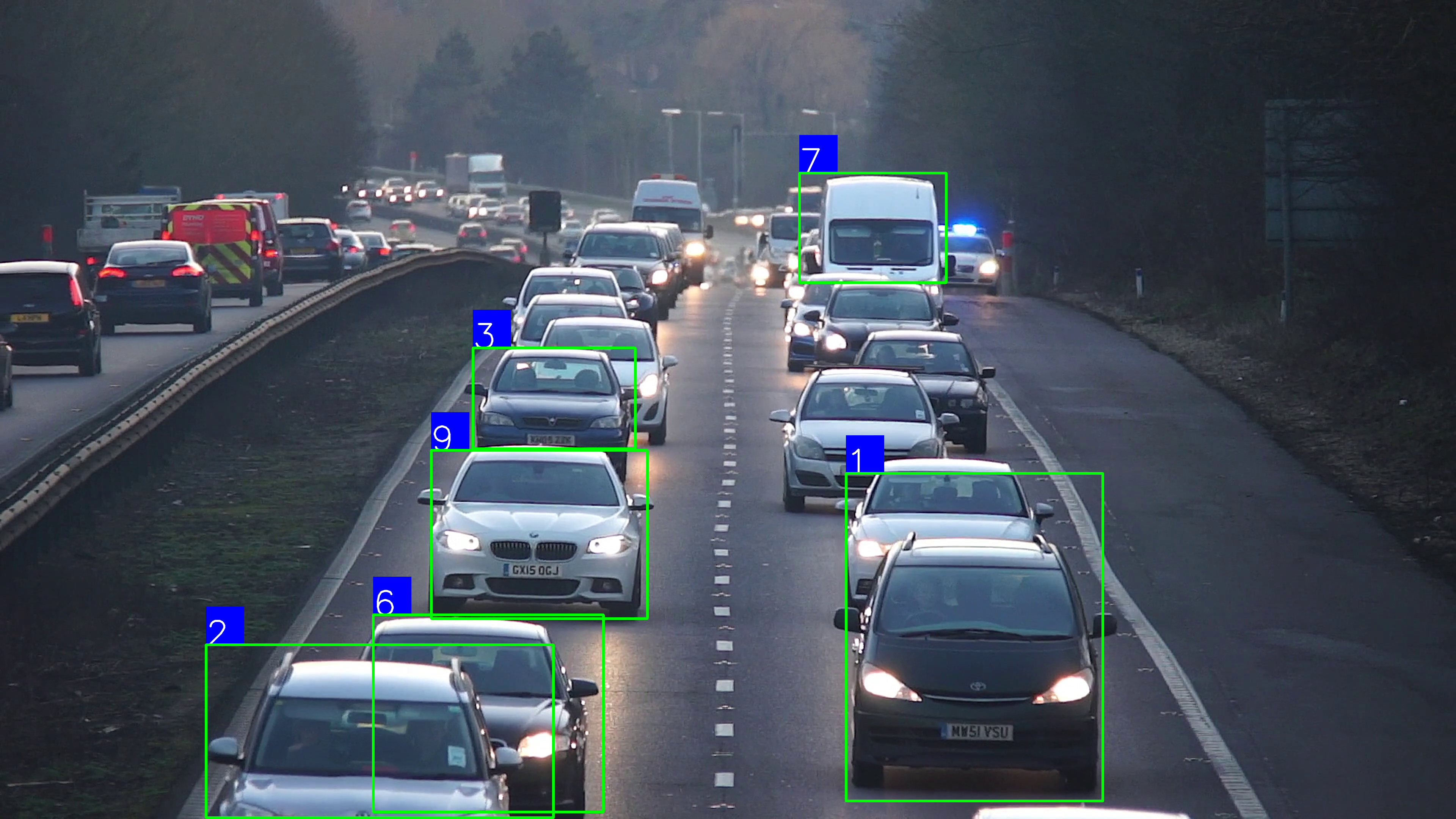}
        \end{subfigure}
    
        \caption{Output of Tracking algorithm over time.}
        \label{fig:three_graphs}
    \end{figure}  
    
    \item State Estimation:
    The Kalman filter \cite{welch1995introduction} is used to estimate the tracked object's future state by fusing the anticipated state based on the dynamics model of the object with fresh sensor inputs. The object's position, velocity, and other pertinent states are improved over time as a result of the iterative updating of the state estimate that takes into account the uncertainty of both the prediction and the measurements. The states are as follows:
    \begin{itemize}
        \item $x_c$: centroid of the bounding box in a column direction.
        \item $y_c$: centroid of the bounding box in a row direction.
        \item $s$: scale or area of bounding box.
        \item $a$: aspect ratio of the bounding box, assume to be constant.
        \item $\dot x_c$: velocity of the centroid ($px/sec$) of the bounding box in a column direction.
        \item $\dot y_c$: velocity of the centroid ($px/sec$) of the bounding box in a row direction.
        \item $\dot s$: velocity of scale or area of bounding box. 
    \end{itemize}
\end{itemize}
    
\subsection{Depth Estimation}
To maintain a fixed clearance between the leader and follower robots or to avoid any dynamic obstacles in the environment, it is essential that the follower robots are spatially aware. Depth estimation adds a new dimension to the perception capabilities of the follower robots, enabling them to have a more comprehensive understanding of their surroundings. Additionally, depth information helps the follower robot navigate along the same path as the leader robot. By estimating the depth of the environment, the follower can align itself with the leader's trajectory and adjust its own movements accordingly. This enables the follower to position itself appropriately and maintain the desired formation.

Depth estimation from a single RGB frame often requires complex algorithms. These algorithms can be computationally intensive, making it challenging to achieve real-time performance on resource-constrained platforms or in applications that require fast processing. Recent developments in deep learning solutions such as MiDaS have achieved notable advancement in monocular depth estimation from a single-view. MiDaS has demonstrated impressive performance on various benchmark datasets and has been utilized in several practical applications. The MiDaS model is adopted to create a depth map for the follower robots. 

\begin{figure}[h!]
    \centering
    \includegraphics[width=0.8\linewidth]{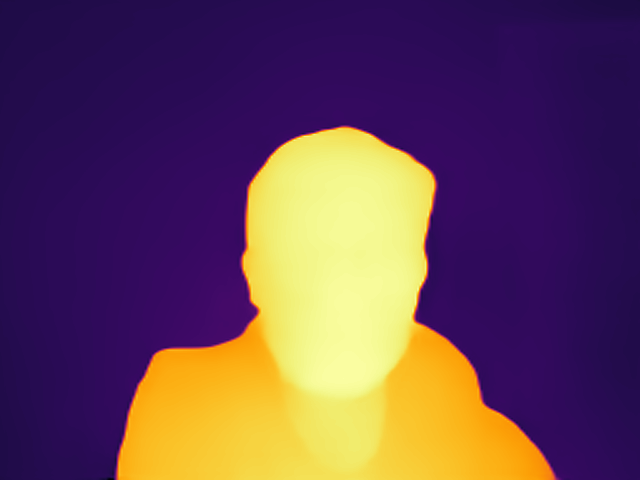}
    \caption{Relative depth map from MiDaS}
    \label{fig:rel_depth}
\end{figure}

The resulting depth estimated by the MiDaS model is relative depth rather than absolute depth (Fig.\ref{fig:rel_depth}). Relative depth refers to the depth information provided in a relative scale, where the depth values are proportional to each other but not directly related to real-world measurements. To obtain actual or metric depth measurements, additional calibration or post-processing steps are required. This step involves using a known reference distance in the scene to scale the relative depth values to real-world measurements. The calibration process is as follows,

\begin{equation*}
    D_{act} = \dfrac{D_{rel}*D_{ref}}{D_{rel\_ref}}
\end{equation*}
where $D_{rel}(x, y)$ is the relative depth value at pixel coordinates (x, y) in the relative depth map, $D_{ref}$ be the known depth value of the reference distance,  $D_{rel\_ref}$ is the relative depth value corresponding to the reference distance and $D_{act}(x, y)$ be the actual depth value at pixel coordinates $(x, y)$ in the calibrated depth map.

The estimated state from the previous step along with the depth information drives the decision-making process for the follower robots. Using the centroid coordinates ($x_c, y_c$) of the bounding box from the state estimation step, the depth information of the tracked entity is obtained from the depth map. It is possible that the centroid of the bounding box may not align perfectly with a single pixel in the depth map. In such cases, the accuracy of the estimated depth value is augmented by averaging the depth values of its neighbouring pixels achieving sub-pixel accuracy. The final depth value is given as follows,

\begin{equation*}
\begin{aligned}
    D_{act} = (1 - \Delta x)(1 - \Delta y)D_{11} + \Delta x(1 - \Delta y)D_{21} \\
    + (1 - \Delta x) \Delta y D_{12} + \Delta x \Delta y D_{22}    
\end{aligned}
\end{equation*}
where $D_{11}, D_{12}, D_{21}, D_{22}$ are the neighbouring pixels and $\Delta x = x_{sub} - x_1 \text{ and } \Delta y = y_{sub} - y_1.$ and ($x_{sub}, y_{sub}$) are the sub-pixel coordinates.

\subsection{Dynamic Planner}
\begin{figure}[h!]
    \centering \includegraphics[width=1\columnwidth]{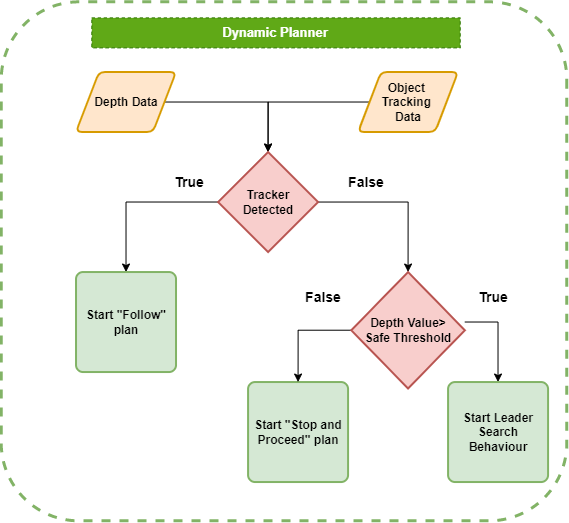}
    \caption{Flow chart of the Dynamic Planner}
    \label{fig:Dynamic_planner}
\end{figure}

The dynamic planner creates the plan for the agents (robots) based on the pose of the target object. The planner creates various plans depending on whether the target object is the leader (an agent to be followed) or the obstacle. When a target object is an obstacle, a "Stop and Proceed" plan is developed, in which the system stops and waits until the obstacle exits the target frame before moving to the goal. Instead, if the planner determines that the target object is the agent to be followed, it develops a "Follow" plan in which the target object is closely followed in order to reduce the angular deviation and linear deviation by a threshold.

As shown in Fig.\ref{fig:Dynamic_planner}, the real-time sensor data from the IMU and the depth data from the Midas depth estimation model are used to estimate the distance and orientation of the follower with respect to the pose of the leader or obstacle.

\subsection{Cooperative Sensing and Communication} 
\begin{figure}[h!]
    \centering \includegraphics[width=1\columnwidth]{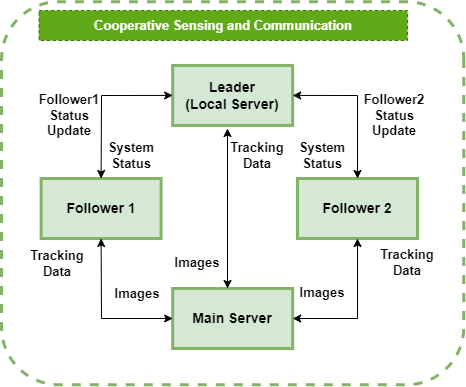}
    \caption{Network architecture for Cooperative Sensing and Communication module}
    \label{fig:CSCM}
\end{figure}

This module is responsible to keep the agents connected with a central knowledge base. It is essential to share important information among the agents in case of an unaccounted event. For instance, in case of an obstacle obstruction being detected, all the agents are informed of the same and until the situation is handled the system goes into a 'stop' state.

The Fig.\ref{fig:CSCM} shows the HTTP protocol-based network architecture supporting the intra-communication framework established between the agents(leader and follower) and the inter-communication framework established between the agents and the main server. The camera feed of each agent is sent to the main server to obtain the model outputs(depth and tracking data) of the perception stack necessary for decision-making. Additionally, the local server kept by the leader is updated in real-time with each follower's status. Each follower simultaneously accesses the server's status of the entire system in order to make real-time decisions.

\subsection{Close-range agent tracking}
\begin{figure}[h!]
    \centering \includegraphics[width=1\columnwidth]{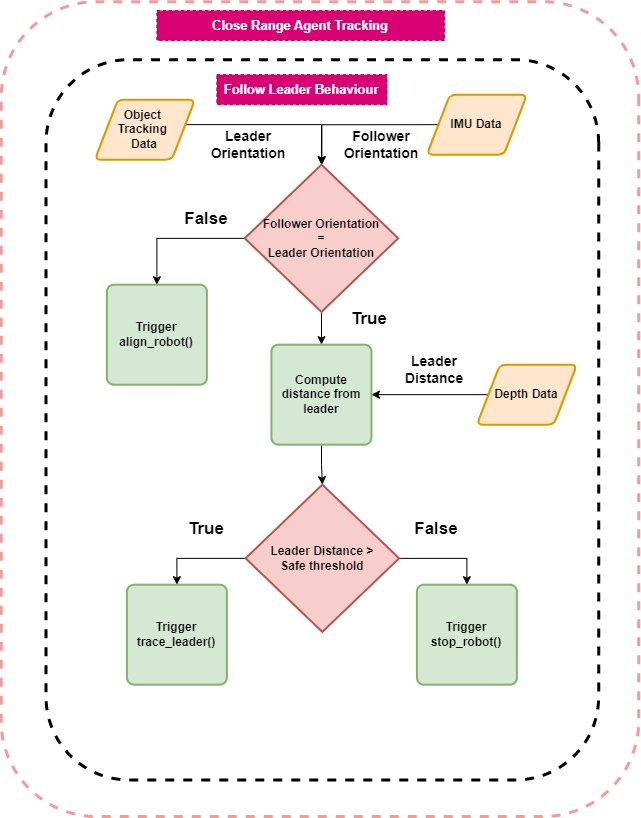}
    \caption{Low-Level Controller architecture of Close Range Agent Tracking Module}
    \label{fig:CRAT}
\end{figure}

The low-level controllers take the motion plan generated by the Dynamic Planner to accurately track and align the target object(an agent to be followed) or avoid the target object(an obstacle) with the help of the feedback from the Inertial Measurement Unit(IMU) and depth data obtained from the Midas model as shown in the Fig.\ref{fig:CRAT}.

\section{Hardware}
\subsection{Setup}
\begin{figure}[h!]
    \centering
    \includegraphics[width=0.8\linewidth]{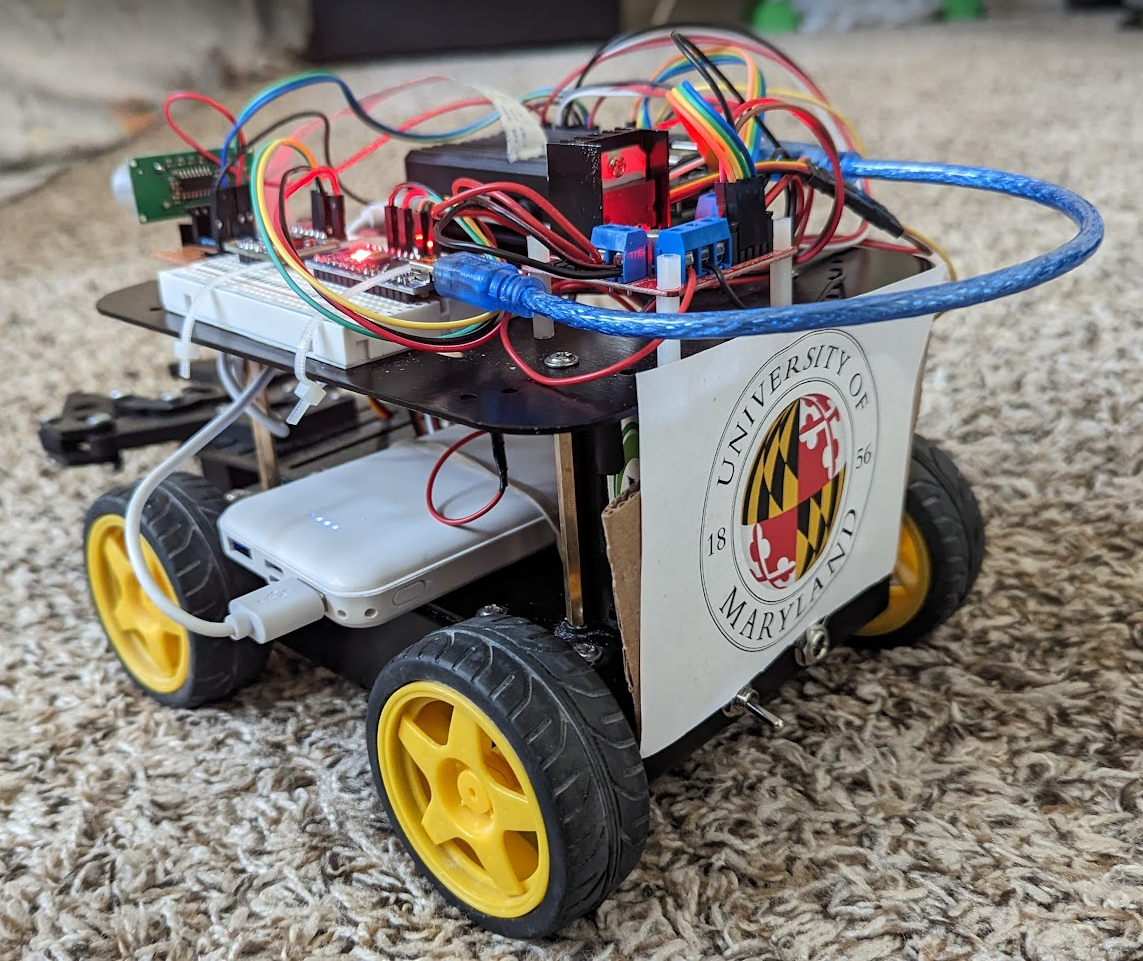}
    \caption{The robot that acted as both the leader and follower}
    \label{fig:Robot}
\end{figure}

The proposed auto platoon system was tested on robots previously built for the Autonomous Robotics course. These robots are powered by Raspberry Pi 3B+ running a 1.4GHz 64-bit quad-core ARM Cortex-A53 CPU with 1GB of RAM. The sensor suite includes a Raspberry Pi camera, an ultrasonic sensor, magnetic encoders for the geared motors and an Inertial Measurement Unit controlled using an Arduino Nano. The robot used for this setup is shown in Fig.\ref{fig:Robot}.
{\color{white} "}

The hardware setup remained the same for both the leader and follower robots. A client-server-based connection was established between the follower robots and a host computer. The lower-level computation which includes processing sensor information and motor control was done on the Raspberry Pi. As the computing resources on the Raspberry Pi do not suffice the requirements to run deep learning models, a host computer powered by NVIDIA RTX 2080 GPU was used to run the depth estimation and state estimation modules.

\subsubsection{Implementation}
The robots were set up so that one of them acted as the leader and the other robot the follower. An image of the trained dataset was added to the rear of the leader robot such that the follower robot could track the target object using the onboard camera, as shown in Fig.\ref{fig:three_graphs}.
{\color{white} "}

The leader robot was given a fixed path to follow and the follower robot autonomously tracks the leader robot while maintaining a safe minimum distance (in this case fixed as 30cm) from it using the depth information of the leader bot. This is done by assigning a unique identification for the initially identified frame.

\begin{figure}[h!]
    \centering
    \includegraphics[width=0.8\linewidth]{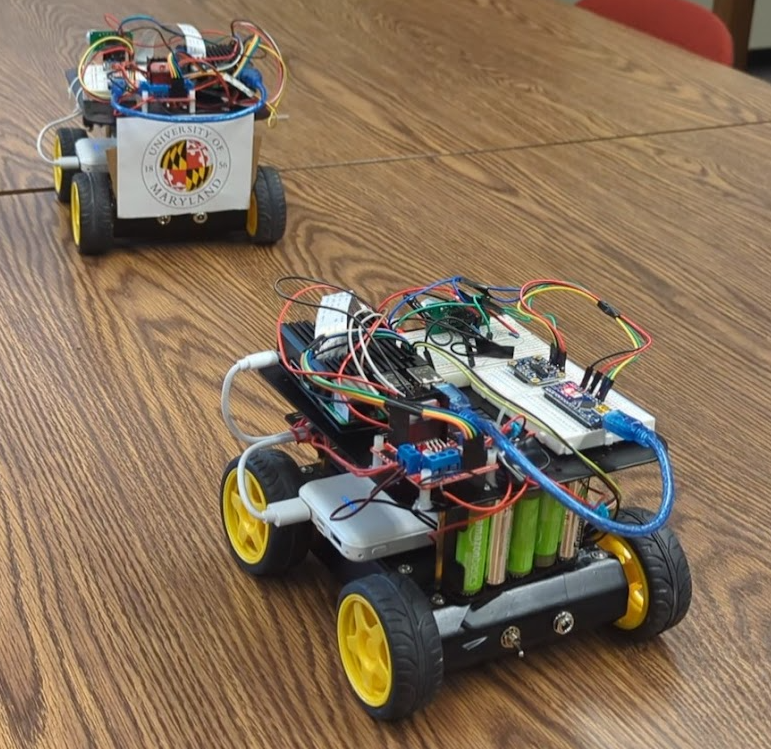}
    \caption{The leader robot and the follower robot}
    \label{fig:three_graphs}
\end{figure}    

\begin{figure}[h!]
    \centering
    \includegraphics[width=0.8\linewidth]{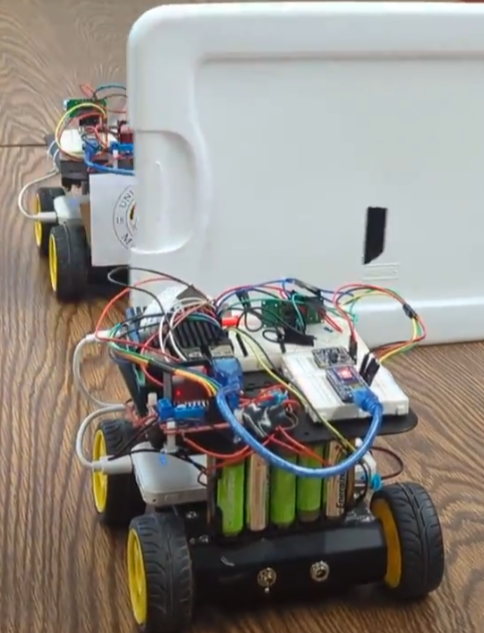}
    \caption{Obstacle in front of the follower}
    \label{fig:obstacle}
\end{figure}

Additionally, dynamic obstacles were introduced between the leader and the follower, and the working of the cooperative sensing and communication network to bring the system to the 'stop' state was tested. When the obstacle was removed it was verified that the follower retains the tracking of the leader robot and the platoon then again resumes the planned path as shown in Fig.\ref{fig:obstacle}.

\section{Results}
The repository containing the code for the work can be found \href{https://github.com/yashveerjain/Leader-Follower}{here}.

\begin{figure}[h!]
    \centering
    \includegraphics[width=\linewidth]{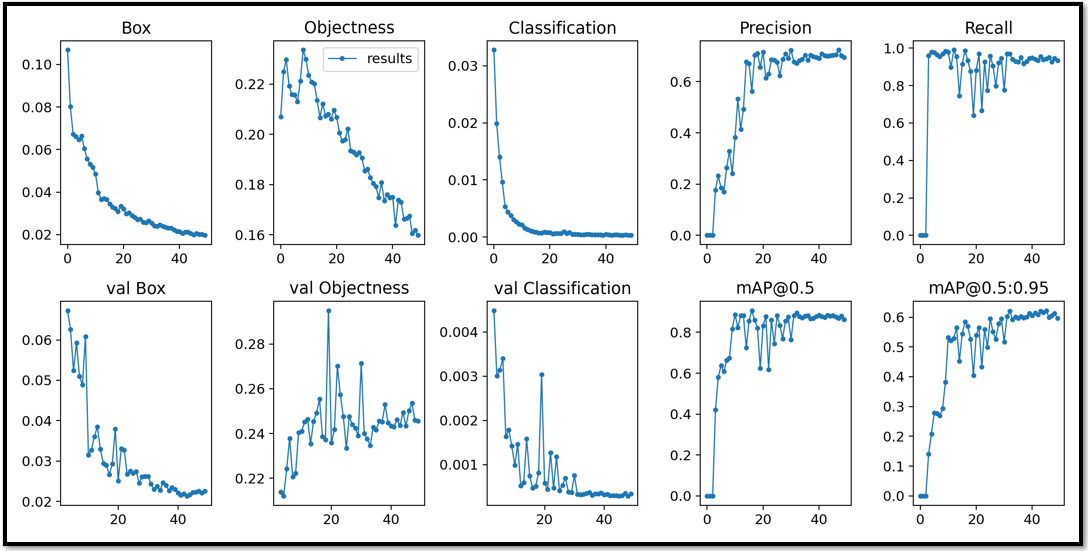}
    \caption{Training details for the Yolov7 model}
    \label{fig:training_dets}
\end{figure}

The results of the object detection phase are outstanding in terms of accuracy and efficiency. The implemented object detection algorithm demonstrated exceptional performance by successfully identifying and localizing objects of interest within the given input data. It effectively detected the target of interest even in complex and occluded backgrounds, with remarkable precision. The results of the model training can be seen in Fig \ref{fig:training_dets}. A higher mAP value indicates better performance, as it signifies that the model has achieved higher precision and recall across the target class. The high mAP value suggests that the model can effectively detect objects with fewer false positives and false negatives. The results of the object detection model on the test images can be seen in Fig \ref{fig:three_graphs}.

\begin{figure}
        \centering
        \begin{subfigure}{0.65\textwidth}
             \includegraphics[width=0.6\textwidth]{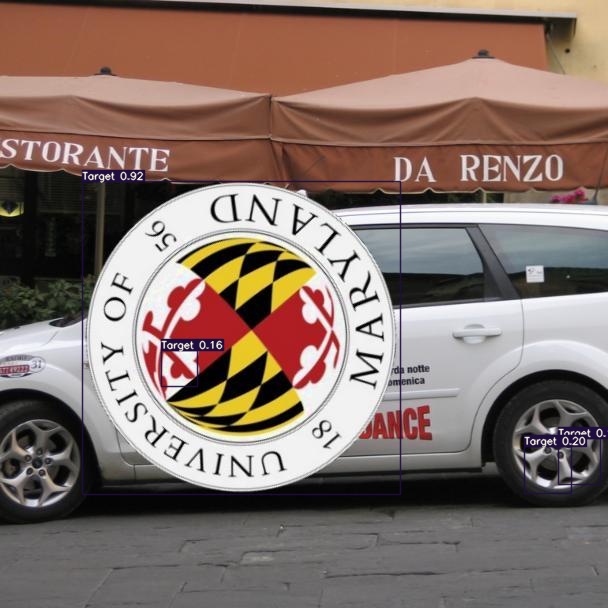}
        \end{subfigure}
        \begin{subfigure}{0.65\textwidth}
             \includegraphics[width=0.6\textwidth]{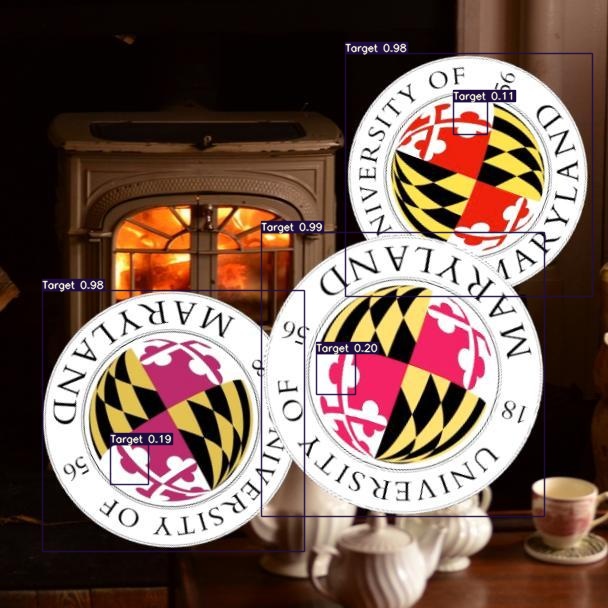}
        \end{subfigure}
        \caption{The results of the detection model on the test data.}
        \label{fig:three_graphs}
\end{figure}  

\begin{figure}[h!]
    \centering
    \includegraphics[width=0.8\linewidth]{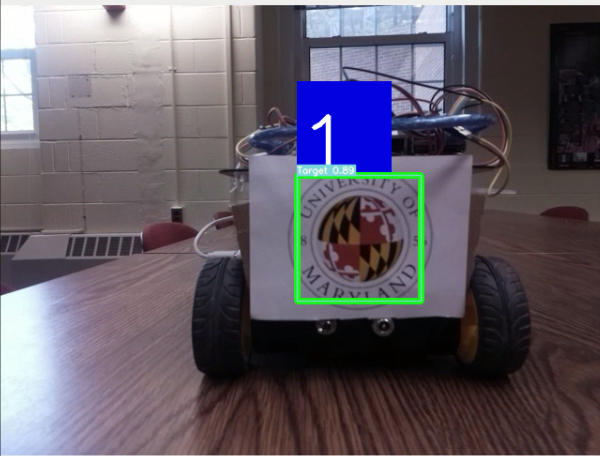}
    \caption{The leader robot is being tracked by the follower robot.}
    \label{fig:tracking_result}
\end{figure}

\begin{figure}[h!]
    \centering
    \includegraphics[width=0.8\linewidth]{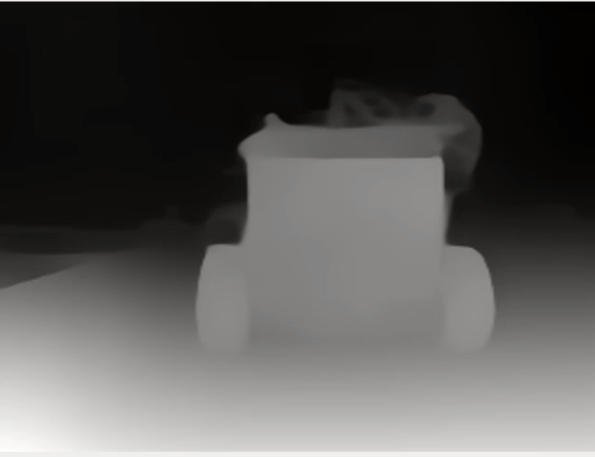}
    \caption{The depth map of the leader robot from the follower's perspective.}
    \label{fig:depth_result}
\end{figure}

In terms of tracking speed, robustness to occlusions, and accuracy the custom tracking algorithm implemented was giving excellent performance. Fig. \ref{fig:tracking_result} shows the real-time result of the tracking algorithm while follower is tracking the leader from behind. As can be seen, the tracked output was given a unique id which remained constant throughout the end of the experiment showing the quality of the model. Similarly, the result of the Midas model as can be seen in Fig.\ref{fig:depth_result} was also highly accurate and met the needs of the implementation.


Demonstration videos for the implemented leader-follower system can be found \href{https://drive.google.com/drive/u/1/folders/1lEYSDns3Q3QMbOsjFyqpdVxPJQgv3-mY} {here}. 

\section{Conclusion}
A novel bio-inspired leader-follower system based on software latching has been presented and examined in this work.  The system pipeline involved a custom object tracking algorithm backed by a trained depth estimation model to extract the real-time pose of the leader(or agent) to be tracked or avoided. Further, a highly comprehensive inter and intra-communication framework was established for efficient communication between the agents of the system and between the agents and the main compute server running the heavy perception models. 

Finally, the efficiency and performance of the implemented pipeline were tested on a custom mobile robot reinforced with a robust low-level controller essential for supporting the motion plan generated by the above software stack. The results of the experiments performed were highly satisfactory.

\section{Future Work}
The future scope of work would be to use an edge device such as NVIDIA Jetson Nano as the onboard computer on the robots. The computation resources afforded by such devices would enable the decision-making and intra-system communication to take place between the participants directly. Additionally, improving the sensor suite of the robots by including a depth camera would drastically increase the efficiency of the system. As this camera would give us ground truth data and the intricacies of estimating the depth through a model would be overcome. Further, the speed of the detected object using the Kalman filter can be used to improve the dynamic obstacle avoidance strategy.

\section{Acknowledgements}
We would like to thank Dr Samer Charifa for giving us this opportunity. Also, we greatly appreciate the efforts put in by the Teaching Assistants of the course ENPM673 for reviewing our submission on time and giving us their valuable feedback. 


{\small
\bibliographystyle{unsrt}
\bibliography{ref.bib}
}

\end{document}